\documentclass{article}

\usepackage{microtype}
\usepackage{graphicx}
\usepackage{subfigure}
\usepackage{booktabs} % for professional tables
\usepackage{amsmath}
\usepackage{hyperref}

\usepackage{hyperref}

\usepackage[accepted]{icml2018}

\icmltitlerunning{Deep Embedding Kernel}

\begin{document}

\twocolumn[
\icmltitle{Deep Embedding Kernel}




\begin{icmlauthorlist}
\icmlauthor{Linh Le}{to}
\icmlauthor{Ying Xie}{goo}
\end{icmlauthorlist}

\icmlaffiliation{to}{Institute of Analytics and Data Science, Kennesaw State University, Georgia, USA
}
\icmlaffiliation{goo}{Department of Information Technology, Kennesaw State University, Georgia, USA
}

\icmlcorrespondingauthor{Linh Le}{lle12@students.kennesaw.edu}
\icmlcorrespondingauthor{Ying Xie}{yxie2@kennesaw.edu}

\icmlkeywords{Machine Learning, ICML}

\vskip 0.3in
]



\printAffiliationsAndNotice{} % leave blank if no need to mention equal contribution
%\printAffiliationsAndNotice%{\icmlEqualContribution} % otherwise use the standard text.

\begin{abstract}
In this paper, we propose a novel supervised learning method that is called Deep Embedding Kernel (DEK). DEK combines the advantages of deep learning and kernel methods in a unified framework. More specifically, DEK is a learnable kernel represented by a newly designed deep architecture. Compared with pre-defined kernels, this kernel can be explicitly trained to map data to an optimized high-level feature space where data may have favorable features toward the application. Compared with typical deep learning using SoftMax or logistic regression as the top layer, DEK is expected to be more generalizable to new data. Experimental results show that DEK has superior performance than typical machine learning methods in identity detection, classification, regression, dimension reduction, and transfer learning. 

\end{abstract}

\section{Introduction}
\label{intro}

We consider two major branches of machine learning, kernel methods \cite{hofmann2008kernel} and deep learning \cite{schmidhuber2015deep}. Kernel methods center around the kernel trick \cite{hofmann2008kernel} -- using a pre-defined kernel function to implicitly map data to a new feature space. However, this implicit mapping is rather heuristic in that there is no guarantee that the pre-defined kernel can lead to a more favorable feature space where data has better distribution towards the application. Hyper-parameter tuning algorithms like grid-search may improve the model performance (i.e. less prediction errors), but this brutal-force strategy does not fundamentally solve the problem of using pre-defined kernels. 

Deep learning, on the other hand, utilizes a high number of parameters structured by layers of neural networks to map the data to an explicit feature space with specified dimensionality \cite{schmidhuber2015deep}. The parameters of the network that determines the mapping are typically tuned based on an explicit learning objective. In other words, by deep learning, the mapping of data into high-level representations is directly guided by the given learning objective through some top-down learning processes such as gradient descend. Therefore, learning objectives play critical roles in the quality of mapping. Frequently used learning objectives try to minimize training errors, which may not have the desired generalization ability according statistical learning theory \cite{vapnik1999overview}. The work in \cite{tang2013deep} tries to improve generalization ability of deep learning by using linear SVM at the top layer, but the computational complexity of integrating SVM to deep learning is high. Another restriction of deep learning is that the dimensionality of the mapped feature space is pre-specified, instead of being learned. 

In this paper, we try to address the problems of both kernel machines and deep learning by proposing a new supervised learning method called Deep Embedding Kernel (DEK) that is able to utilize the strengths of each method to address the weakness of the other in a unified framework. First of all, DEK does not explicitly map data to a feature space with pre-specified dimensionality, nor implicitly map data through a pre-defined kernel; instead, DEK uses a newly designed deep architecture to represent a learnable kernel. In other words, DEK utilizes the learning power of deep learning to train a kernel, which in turn implicitly maps data to a high dimensional feature space. The learning objective of DEK specifies a desired relationship of data in the mapped feature space. Then the kernel represented by DEK trained by the learning objective is expected to implicitly map data to such a feature space. Therefore, the whole mapped feature space, including its dimensionality, is learned via deep learning. Using deep architectures to learn a kernel, instead of directly learn the feature space also has the advantages of flexibility in that the learned kernel can be applied to a wide range of supervised learning tasks including identity detection, general classification, dimension reduction, regression, and other kernel based machine learning applications.

The architecture of DEK integrates two learning networks, namely \textbf{kernel network} and \textbf{embedding network}. The kernel network directly represents the parameterized kernel trained from data, while the embedding network tries to learn optimized data representations to feed into the kernel network. The training of both networks is done in a single gradient descent process with the same learning objective that specifies an optimized relationship of data in the desired feature space. 

DEK can be easily extended to work on unstructured data by laying itself on top of deep architectures designed for certain type of unstructured data, such as Convolutional Neural Network (CNN) for image data, Recurrent Neural Network (RNN) for sequential data, or the combination of CNN and RNN for video data. By this extension, the particular deep architecture used on unstructured data will learn vector embedding from unstructured data in the same learning process where embedding network and kernel network of DEK are trained via gradient descent. Moreover, DEK can be used to boost the learning power of transfer learning by being laid over a trained deep network that outputs vector embedding. 

In this paper, we will demonstrate that DEK has superior performance over other typical supervised learning methods, such as Kernel Support Vector Machines, Gradient Boosting Trees, Random Forests, and Neural Networks on multiple learning tasks, including identity detection, general classification, regression, dimension reduction, and transfer learning. 

\section{Related Works}
Various attempts were made to stack kernels to form deep architectures in \cite{zhuang2011two}, \cite{strobl2013deep}, \cite{jose2013local}, \cite{jiu2017nonlinear}, and \cite{sahbi2017coarse}. The output of this type of deep architecture is typically a highly nonlinear combination of input kernels. The learning process of stacking kernels involves jointly training a SVM classifier and modifying network weights as well as kernel parameters using gradient descent. Some limitations of these works include 1) using pre-defined kernels (such as RBF kernel) as input neurons limits the flexibility and capacity of learning by the deep architecture; 2) using SVM optimization as the learning objective for training the deep architecture is computationally expensive. The proposed DEK tries to maximize the learning by first learning an optimal high-level representation of data, followed by learning a highly non-linear kernel, which is in turn based on dimension-wise relationships of the high-level representations. In other words, DEK forms the kernel based on much finer granularity of relationship between data instead of starting with pre-defined kernel functions on the whole set or subsets of dimensions of data. Furthermore, the learning objective of DEK can be evaluated online by each pair of data without the need of quadratic programming on at least a batch of data. 

Similarly, stacking SVMs to deepen the model architecture was discussed in \cite{wiering2014multi}. The authors of this work use different SVMs to extract latent features in different subsets of dimensions in the data. A global SVM is then used to aggregate all SVMs to form a final decision layer. However, because of computational expenses of SVMs, it is not practical to form a deep architecture by simply stacking SVMs. Therefore, the extent to which this type of stacking takes advantages of deep learning is rather limited. On the contrary, DEK can fully embrace the learning power of deep learning, given that DEK itself is a true deep architecture without any add-on restriction on depths of the network. Instead of stacking SVMs, the work in \cite{tang2013deep} tried to improve generalization ability of deep learning by using linear SVM classifier at the top layer to define the learning objective. But this architecture strictly ties with classification tasks and training a SVM at the top layer is still non-trivial as it requires quadratic programming on a batch of data. 

There were works computing similarity of data using deep architectures on image data in \cite{zbontar2015computing} and \cite{zagoruyko2015learning}. However, their similarity computing is specialized on a particular task and unable to be generalized to other learning tasks. Moreover, their output similarities do not necessarily possess the character of symmetricity, therefore cannot be used as kernels. 

Google’s FaceNet uses a cost function that is called triplet loss on facial identification \cite{schroff2015facenet}. Each evaluation of triplet loss involves selecting three instances ${x_i, x_i^+, x_i^-}$ that satisfies the following criteria: $x_i$ is an anchor point, $x_i^+$ is another data point with the same class as $x_i$, $x_i^-$ is a data point with a different class than $x_i$, and the following inequality holds.

\begin{equation}
\Vert x_i - x_i^+ \Vert_2^2 > \Vert x_i - x_i^- \Vert_2^2
\end{equation}

The deep network is then trying to learn an mapping $f(\cdot)$ such that

\begin{equation}
\Vert f(x_i) - f(x_i^+) \Vert _2^2 < \Vert f(x_i) - f(x_i^-) \Vert_2^2 \qquad \forall i
\end{equation}

Therefore, the learning objective of the deep learning can be expressed as minimize the following cost function:

\begin{equation}
L = \sum_{i=1}^N(\Vert f(x_i ) - f(x_i^+ ) \Vert_2^2 - \Vert f(x_i) - f(x_i^-)\Vert_2^2 + \alpha)
\end{equation}

with $\alpha$ being a margin parameter. The Triplet Loss function was extended to other identity detection tasks such as voice recognition \cite{bredin2017tristounet}. An issue with triplet loss based cost function, according to \cite{hermans2017defense}, is that the training of the network requires a large training data that contains a sufficient amount of triplets that satisfies the described criteria. In contrast, DEK can evaluate the learning object online by using every pair from the training data (though it is not necessary to use every pair if the training data is large enough). From another perspective, DEK may even be able to solve the "Small Training Data" problem by forming $n^2$ training pairs from just $n$ instances. 

Lastly, we would like to mention transfer learning. In the context of deep learning, transfer learning aims to reuse a deep network that is trained for one application to another relevant task \cite{pan2010survey} and \cite{bengio2012deep}. A popular way of doing transfer learning is to replace the decision layer(s) of the trained deep network with a new one for the new task. DEK can work as a general decision layers to be laid on top of a pre-trained network. Experimental results demonstrate that DEK has better performance than Multi-layer Perceptron with triplet loss for being used as the decision layers in transfer learning. 

\section{Methodology}

\begin{figure}[tb]
\vskip 0.2in
\begin{center}
\centerline{\includegraphics[width=\columnwidth]{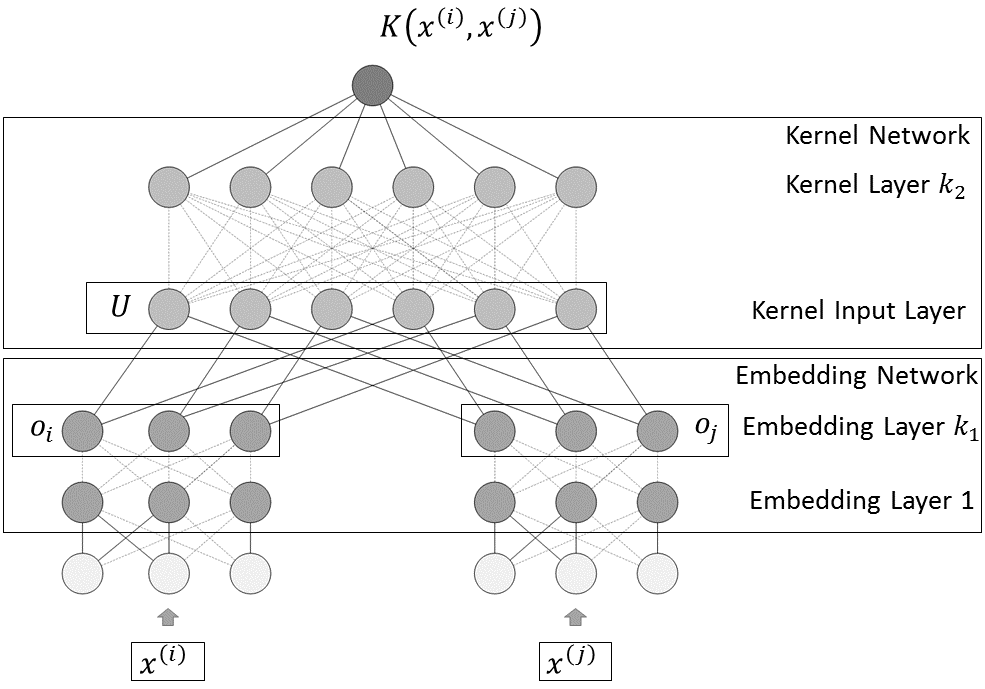}}
\caption{The Structure of DEK}
\label{DEKstructure}
\end{center}
\vskip -0.2in
\end{figure}

The goal of our methodology is to learn an optimized feature space of data with desired features for the application. This optimized space is determined by DEK, a learnable kernel that is represented by a deep architecture. When we design DEK, we consider the following factors. First, since it represents a kernel, DEK takes a pair of data instances as input and output their similarity. Similarity of data can be computed based on different representations of data at different abstraction levels. We want the DEK to be able to learn data similarity based on optimized data representations. Then based on the given data representation, we want the DEK to be able to learn a similarity function that is complex enough to map data to an optimized space with desired data distributions. Therefore, DEK is designed to have two learning components, namely embedding network and kernel network, integrated in a unified deep architecture. These two learning components will be trained using the same learning objective in a single learning process. The overall architecture of DEK is shown in Figure \ref{DEKstructure}.

\subsection{Kernel Network}

As shown in Figure \ref{DEKstructure}, the input of the kernel network is denoted as $U$, which is formed by the outputs of the two branches of the embedding network, which are $o_i$ and $o_j$ respectively. More specifically, $U$ can be expressed as the following function of $o_i$ and $o_j$

\begin{equation}
U = 
\left \{
\begin{array}{c}
o_{i_1}*o_{j_1},o_{i_2}*o_{j_2},\dots o_{i_d}*o_{j_d},\\
|o_{i_1}-o_{j_1}|,\dots |o_{i_d}-o_{j_d}|
\end{array}
\right \}
\end{equation}

Where $o_{i_k}$ denotes the $k^{th}$ dimension of $o_i$, and $d$ is the dimensionality of $o_i$ and $o_j$ . In other words, each neuron in the input layer of the kernel network represents a symmetric relationship of $o_i$ and $o_j$ on a single dimension. The use of fine granularity of relationship on each individual dimension as input provides more room for learning, compared with directly using different pre-defined kernel functions on $o_i$ and $o_j$ as inputs. Essentially, this design of inputs allows the kernel network to learn a kernel that is a highly nonlinear combination of angles and distances of the data pairs in the space that is learned by the underneath embedding network. Furthermore, this design of inputs guarantees the output similarity is symmetric. 

The output of the kernel network is the probability that sample $i$ and $j$ belong to the same class. Formally, given sample $i$ and $j$, the output can be expressed as 

\begin{equation}
\begin{aligned}
K(o_i,o_j ) = P(y^{(i)}=y^{(j)} | x^{(i)},x^{(j)}) \\
= sigmoid(W_{out}^{(K)} \cdot H_{out}^{(K)} + b_{out}^{(K)})
\end{aligned}
\end{equation}

With $W_{out}^{(K)}$ and $b_{out}^{(K)}$ being the parameters of the output layer, and $H_{out}^{(K)}$ being the input, of the kernel network. Obviously we have $K(o_i,o_j) = K(o_j,o_i) > 0$. Therefore, $K(\cdot)$ is a kernel function. 

To train the kernel network (as well as the whole DEK), we label each pair in the following way. 

\begin{equation}
\begin{cases}
Y^{(i,j)} = 1 \iff y^{(i)} = y^{(j)} \\
Y^{(i,j)} = 0 \iff y^{(i)} \neq y^{(j)}
\end{cases}
\end{equation}

That is, if instance $i$ and $j$ belong to the same class, the label for the pair of $i$ and $j$ is 1, otherwise it is 0. Then we define the learning objective of training DEK (including kernel network) is to minimize the following cost function. 

\begin{equation} \label{dekcost}
\begin{aligned}
L = \sum_{data}(Y^{(i,j)}\log(K(o_i,o_j) + \\
(1-Y^{(i,j)})\log(1-K(o_i,o_j))) 
\end{aligned}
\end{equation}

\subsection{Embedding Network}
The purpose of the embedding network is to learn optimized high-level representations of data to feed into the kernel network as inputs. Let the mapping made by the embedding network be $E(\cdot)$ then the high-level representation of sample $x^{(i)}$ can be represented as $o_i=E(x^{(i)})$. The goal of designing the embedding network is to increase the learning capacity of the final kernel. Experimental results demonstrate that the embedding network positively contributes to the performance of DEK. 

The training of embedding network is in the same gradient descent process using the same cost function as in Equation (\ref{dekcost}). 

\subsection{Overall Design}
Suppose the embedding network has $k_1$ hidden layers $H_1^{(e)} \dots H_{k_1}^{(e) }$ and the kernel network has $k_2$ hidden layers $H_1^{(K)} \dots H_{k_2}^{(K)}$. Also suppose the input layer of the embedding network is $H_0^{(e)}$ and of the kernel network is $H_0^{(K)}$, and the weights and bias of layer $i$ of network $j$ are $W_i^{(j)}$ and $b_i^{(j)}$. The computational flow from a sample pair $(x^{(i) },x^{(j)})$ can be expressed as

\begin{itemize}
\item The embedding of $x^{(i)}$:

$H_0^{(e)}(i)=x^{(i)}$

$H_1^{(e)}(i)=\sigma(W_0^{(e)}\cdot H_0^{(e)}(i) + b_0^{(e)})$

\dots\\

$o_i=H_{k_1}^{(e)} (i)=\sigma(W_{k_1-1}^{(e)}∙H_{k_1-1}^{(e)} (i)+b_{k_1-1}^{(e)})$

\item The embedding of $x^{(j)}$:

$H_0^{(e)}(j)=x^{(j)}$

$H_1^{(e)}(j)=\sigma(W_0^{(e)}\cdot H_0^{(e)}(j) + b_0^{(e)})$

\dots\\

$o_j=H_{k_1}^{(e)} (j)=\sigma(W_{k_1-1}^{(e)}∙H_{k_1-1}^{(e)} (j)+b_{k_1-1}^{(e)})$

\item Input to the kernel network:

$U=o_i \bullet o_j = H_0^{(K)}$

$H_1^{(K)}=\sigma(W_0^{(K)}\cdot H_0^{(K)} + b_0^{(K)})$

\dots\\

$H_{k_2}^{(K)}=\sigma(W_{k_2-1}^{(K)}\cdot H_{k_2-1}^{(K)} + b_{k_2-1}^{(K)})$

$K(x^{(i)},x^{(j)})=s(W_{k_2}^{(K)}\cdot H_{k_2}^{(K)}+b_{k_2}^{(K)})$

\end{itemize}

with $\sigma(\cdot)$ being the activation function, $s(\cdot)$ being the output function, and $\bullet$ being the dimension-wise similarity operator as discussed:

\begin{center}
$o_i \bullet o_j = 
\left \{
\begin{array}{c}
o_{i_1}*o_{j_1},o_{i_2}*o_{j_2},\dots o_{i_d}*o_{j_d},\\
|o_{i_1}-o_{j_1}|,\dots |o_{i_d}-o_{j_d}|
\end{array}
\right \}
$
\end{center}

Layers in both component network are updated with gradient descent:

\begin{equation}
\begin{aligned}
W_i^{(j)} \leftarrow W_i^{(j)} - \alpha \frac{\partial L}{\partial W_i^{(j)}}\\
b_i^{(j)} \leftarrow b_i^{(j)} - \alpha \frac{\partial L}{\partial b_i^{(j)}}
\end{aligned}
\end{equation}

A unified structure is currently being employed on all layers to simplify the training process. In detail, all embedding layers have $k$ hidden neurons, and all kernel layers have $2k$ neurons, where $k=\alpha d$ with $d$ being the dimensionality of the original data and $\alpha$ being an integer factor (typically, we use $\alpha \in \{1,2,3,4\}$). 

\section{DEK for Unstructured Data}
\begin{figure}[bt]
\vskip 0.2in
\begin{center}
\centerline{\includegraphics[width=\columnwidth]{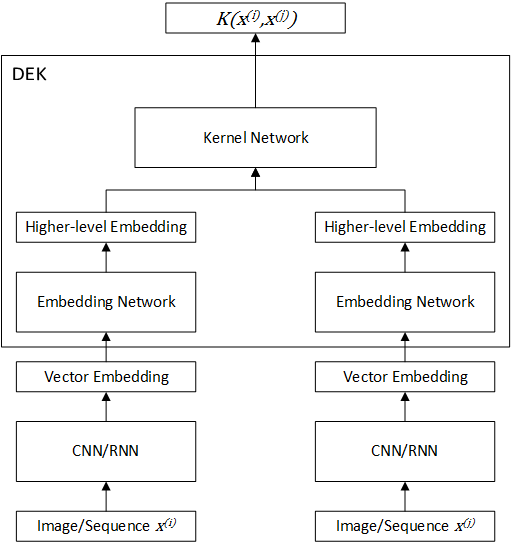}}
\caption{The DEK Architecture for Unstructured Data}
\label{DEKforUData}
\end{center}
\vskip -0.2in
\end{figure}

If data is not in the form of structured records (such as images, sequential data, or text), we can lay DEK on top of CNN, RNN, or other deep architectures that are suitable for the given unstructured data to form a unified deep neural network for supervised learning. The deep neural network with DEK on top for both image data and sequential data are shown in Figure \ref{DEKforUData}. 

In this type of deep architecture, there are three integrated learning components that will be trained in the same learning process. The first is to learn an optimized vector embedding of the unstructured data; the second is to learn an optimized high-level embedding based on the bottom vector embedding; and the last is to learn a complex similarity function based on the high-level embedding of the data. Again, all these components will be trained in the same learning process with the same learning objective.

The deep architecture, shown in \ref{DEKforUData}, can be viewed as a framework for transfer learning. In other words, the bottom component of vector embedding can be replaced by a network that is trained from data with similar natures. 

\section{DEK for Different Types of Supervised Learning}
In this section, we describe different applications of DEK on supervised learning including identity detection, classification, regression, dimension reduction, and transfer learning. All experiments for each of the above tasks are conducted in Python version 2.7.12. Deep models are implemented using the package Theano \cite{bergstra2010theano}, other machine learning models (including the regular MLP) are from the Sci-Kit Learn \cite{pedregosa2011scikit} package. Visualizations are generated using the Matplotlib library \cite{hunter2007matplotlib}. 

\subsection{Identity Detection}
The problem of identity detection can be defined as assigning an identity to a query sample (e.g. a speech segment or a facial image). A common supervised learning strategy to solve this problem is to assign the identity to the query sample based on its nearest neighbors in the training set. Identity detection with DEK feeds the query sample and each of the training sample into the trained deep network and finds the nearest neighbors of the query sample using the outputted kernel values. In our experimental studies, we apply DEK to both speaker identification and facial recognition. Since both tasks are based on unstructured data (i.e. speech segments and facial images), we use the extended DEK framework discussed in section 4. In other words, we lay DEK on top of deep architectures that are proper to underneath unstructured data.

\textbf{Speaker Identification}

Most speaker identification models work by first extracting features from the speech segments. We choose spectrograms as the feature set to be modeled in this task. In short, spectrograms are representations of audio segments in the time/frequency space and have characteristics similar to images. Therefore, CNN is a proper deep architecture to model spectrograms. In other words, we lay DEK on the top of CNN to model the similarities of the speech segments. 
 
In our experiment on speaker identification, we use the Characterizing Individual Speakers (CHAINS) dataset - \cite{cummins2006chains}. The data consists of speech segments from 36 persons in various speaking conditions. In our study, we use only segments recorded in studio where the speakers read scripts in normal talking speed. In the preprocessing phase, we first split the speech segments into syllables using silent gaps; then pad each of them to be 1.5-second-long; and finally transform them into spectrograms. We compare two models to identify the speakers, one is a CNN using Triplet Loss cost function (CNN/TL), and the other is extended DEK laid on top of the same CNN. The preprocessed data is split into 75\% training and 25\% testing.

\begin{figure}[bt]
\vskip 0.2in
\begin{center}
\centerline{\includegraphics[width=\columnwidth]{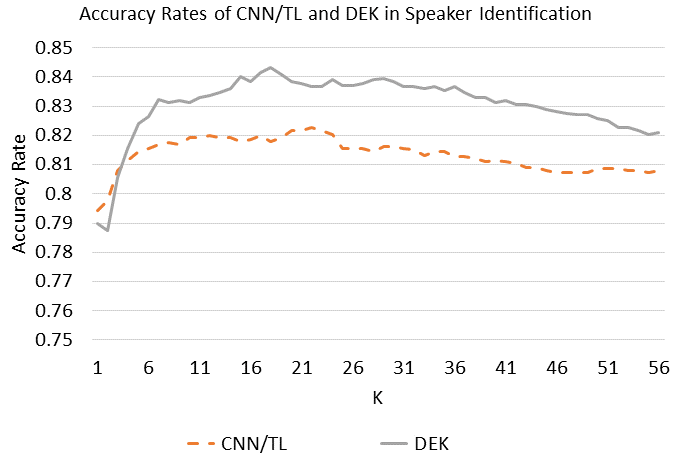}}
\caption{Accuracy Rates of Speaker Identification Models}
\label{speakerModels}
\end{center}
\vskip -0.2in
\end{figure}

The accuracy rates of the two models by number of nearest neighbors from 1 to 55 is shown in Figure \ref{speakerModels}. It can be seen that the DEK provides a significant lift in accuracy rate (over 2\%) over the CNN/TL model.

\textbf{Facial Recognition}

We study the performance of DEK on transfer learning on facial recognition. The data we use is Indian Movie Face Database (IMFDB) -- \cite{setty2013indian}. This dataset contains facial images of Indian movie actors and actresses. We build two transfer learning models based on a pretrained Google FaceNet (available from \small{\url{https://github.com/davidsandberg/facenet}}). \normalsize This version of FaceNet was trained from about 500,000 facial images. We build two transfer learning models based on the pretrained Facenet. One model lays a Multi-layer Perceptron using Triplet Loss (MLP/TL) on top of the pretrained FaceNet, the other lays DEK on the pretrained Facenet. Both models are trained and tested on the same subsets from the IMFDB data (75\% training, 25\% testing). The trained MLP/TL outputs vector embedding based on which we can compute the pairwise distances among images. The trained DEK outputs kernel values that can be interpreted as similarities.

\begin{figure}[bt]
\vskip 0.2in
\begin{center}
\centerline{\includegraphics[width=\columnwidth]{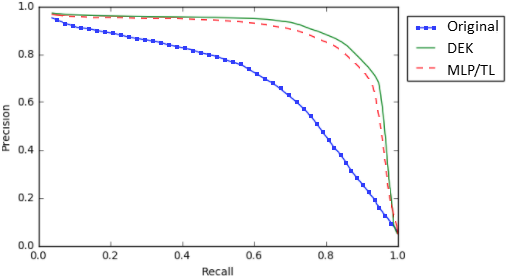}}
\caption{Performance of Transfer Learning Models with DEK and MLP/TL on Facial Recognition}
\label{facialModels}
\end{center}
\vskip -0.2in
\end{figure}

To evaluate the two models, each image in the testing set is used as a query image to rank all images in the training set in the ascending order of their distances outputted by the MLP/TL model, and in the descending order of similarities outputted by DEK. We then plot the average precision-recall curve for these two rankings. We also plot the precision-recall curve generated by the pretrained FaceNet without transfer learning as the baseline. As shown in Figure \ref{facialModels}, both transfer learning models make substantial improvements over the pretrained FaceNet. DEK makes further improvement over MLP/TL at almost every recall level. Given MLP/TL has already achieved near-perfect precisions, the further improvement made by DEK is significant. Therefore, DEK can be used as the desired solution to facial recognition in critical applications where very high accuracy is demanded.

\begin{figure}[bt]
\vskip 0.2in
\begin{center}
\centerline{\includegraphics[width=\columnwidth]{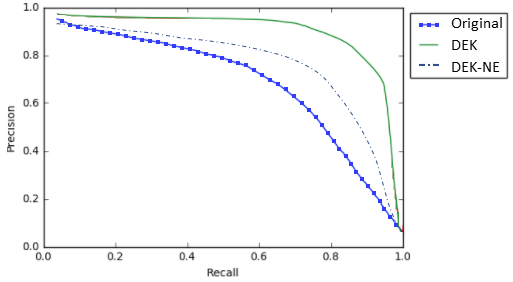}}
\caption{Contributions of Embedding Layers to DEK Performance}
\label{facialModels2}
\end{center}
\vskip -0.2in
\end{figure}

We further study the contribution of the embedding network of DEK towards the performance in this experiment. More specifically, we build a transfer learning model by only laying the kernel network component of DEK on the top the pretrained FaceNet. We denote this model as DEK-EN. Both DEK and DEK-EN are trained independently on IMFDB. The precision-recall curves of both models are plotted in Figure \ref{facialModels2}. It can be seen that the embedding network of DEK contribute significantly towards the performance. This experimental result re-enforces our hypothesis that the incorporating of the embedding network in DEK increases the learning capacity of the model.

\subsection{General Classification}

\begin{table*}[ht]
\caption{Accuracy Rates of Models in Experiments for Classification}
\label{classtable}
\vskip 0.15in
\begin{center}
\begin{scriptsize}
\begin{sc}
\begin{tabular*}{\textwidth}{c @{\extracolsep{\fill}} ccccccc}
\toprule
Dataset & SVM/\textbf{DEK} & KNN/\textbf{DEK} & SVM/RBF & GB & RF & MLP \\
\midrule
Segment \cite{zhang1992selecting} & \textbf{0.9691} & \textit{0.9678} & 0.9647 & 0.9604 & 0.9610 & 0.9593 \\
Cardiotocography \cite{ayres2000sisporto}& \textit{0.9893} & \textbf{0.9899} & 0.9879 & 0.9825 & 0.9846 & 0.9850 \\
Messidor Features \cite{decenciere2014feedback}& \textbf{0.7803} & \textit{0.7746} & 0.7543 & 0.7110 & 0.7168 & 0.7222 \\
Waveform \cite{breiman1984classification} & \textit{0.8696} & \textbf{0.8704} & 0.8684 & 0.8488 & 0.8456 & 0.8672 \\
Pima Diabete \cite{smith1988using} & \textit{0.7839} & \textbf{0.7865} & 0.7708 & 0.7396 & 0.7604 & 0.7630 \\
\bottomrule
\end{tabular*}
\end{sc}
\end{scriptsize}
\end{center}
\vskip -0.1in
\end{table*}

\begin{table*}
\caption{$R^2$ of Models in Experiments for Regression}
\label{regtable}
\vskip 0.15in
\begin{center}
\begin{scriptsize}
\begin{sc}
\begin{tabular*}{\textwidth}{c @{\extracolsep{\fill}} ccccccc}
\toprule
Dataset & SVR/\textbf{DEK} & KNN/\textbf{DEK} & SVR/RBF & GB & RF & MLP \\
\midrule
Concrete \cite{yeh1998modeling} & 0.8651 & 0.8980 & 0.8702 & \textbf{0.9067} & 0.8751 & 0.8119 \\
Airfoil \cite{brooks1989airfoil} & 0.8242 & \textbf{0.9195} & 0.8371 & 0.8840 & 0.9047 & 0.8568 \\
Energy Efficiency \cite{tsanas2012accurate}& 0.9685 & \textbf{0.9783} & 0.9621 & 0.9775 & 0.9756 & 0.9470 \\
\bottomrule
\end{tabular*}
\end{sc}
\end{scriptsize}
\end{center}
\vskip -0.1in
\end{table*}

\begin{figure}[bt]
\vskip 0.2in
\begin{center}
\centerline{\includegraphics[scale=0.4]{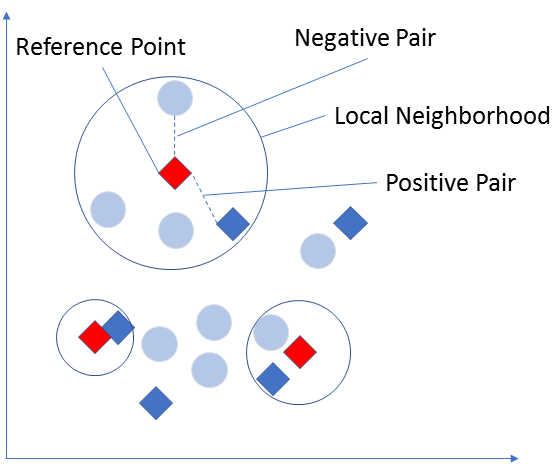}}
\caption{Illustration of the Local Pairing Strategy}
\label{LocalLearning}
\end{center}
\vskip -0.2in
\end{figure}

The learning objective of DEK (described in section 3.1) naturally fits into identity detection problems, in that the desired similarity of two samples belonging to the same identity is 1 and the desired similarity of two samples belonging to different identities is 0. However, for general classification problems, this learning objective may be over-strict, given that two samples belong to the same class may not necessarily have the same level of similarity as two belonging to the same identity. Therefore, to adapt DEK to general classification problems, a local pairing strategy is proposed and added to the learning process of DEK. In details, we use local pairing strategy to generate training pairs at certain interval of iterations. For example, local pairing strategy is applied to generate training pairs at the $1^{st}$, $51^{st}$, $101^{st}$, $151^{st}$, \dots iterations. Other iterations between the interval use the same training pairs generated most recently. The local pairing strategy works as follows. First, all pairs of data are fed into DEK; each sample is used as reference to rank all other samples in descending order of kernel values outputted by DEK. A certain recall level (e.g., 0.1) is then used to determine the neighborhood of the reference sample. Within the neighborhood, we form positive pairs between the reference sample and the samples of the same class, and form negative pairs between the reference sample and the samples of different classes. The local pairing strategy is illustrated in Figure \ref{LocalLearning}. By using local pairing strategy, we avoid to force the similarity of distant samples of the same class to be close to 1.

To study the performance of DEK with local pairing strategy on general classification, we compare SVM using DEK (SVM/DEK) and KNN using DEK (KNN/DEK) with other classification models including SVM using RBF kernel (SVM/RBF), Gradient Boosting Trees (GB) \cite{friedman2002stochastic}, Random Forest (RF) \cite{liaw2002classification}, and MLP on five datasets. The datasets are split into 50\% training data and 50\% testing data. The hyper-parameters used by RBF kernel used by SVM are optimized via grid-search on the trained dataset. Reported accuracy rates are computed in the testing set. Table \ref{classtable} shows the test accuracy rates for all the models.

As can be seen, DEK-based SVM and DEK-based KNN achieve the best results in all datasets. The improvement is from 0.2\% in the Cardiotocography data (comparing to SVM/RBF) to about 7\% in the Messidor Features data (comparing to GB).

\subsection{Regression}

Unlike identity detection or classification models, determining the similarity of a sample pair in regression is not that obvious since the target value is now continuous. When applying DEK to regression, we model the similarity of sample pairs based on the similarity between their target values. In other words, let $K'(\cdot)$ be a similarity function defined on a pair of target values, then the network is trained so that $K(x^{(i)},x^{(j)})$ approximates $K'(y^{(i) },y^{(j)})$:

\begin{equation}
K(x^{(i)},x^{(j)}) \approx K'(y^{(i)},y^{(j)})
\end{equation}

We define $K'(y^{(i)},y^{(j)}) = \exp⁡(-\gamma |y^{(i)}-y^{(j)}|)$ with $\gamma$ being a scale parameter. Accordingly, the output layer and cost function of the regression DEK are

\begin{equation}
K(x^{(i)},x^{(j)}) = ReLU(W_{k_2}^{(K)} \cdot H_{k_2}^{(K)} + b_{k_2}^{(K)} )
\end{equation}
\begin{equation}
L=\frac{1}{N} \sum_{i,j \in Data} (K(x^{(i)},x^{(j)}) - K'(y^{(i)},y^{(j)} ))^2 
\end{equation}

(with $ReLU(x)=\max⁡(0,x)$)

To study the performance of DEK on regression, we compare Support Vector Regressor using DEK (SVR/DEK) and KNN using DEK (KNN/DEK) with other regression models including SVR using RBF kernel (hyper-parameters optimized via grid-search) (SVR/RBF), Gradient Boosting Trees Regressor (GB), Random Forest Regressor (RF), and MLP Regressor, on three datasets. A ratio of 50\% training data and 50\% testing data is also used. $R^2$ is used as the measurement to compare the models. Table \ref{regtable} shows the performances of tested models in the regression task.

As can be seen, KNN/DEK achieves the best performance in two out of three datasets while being slightly behind the GB model in the Concrete dataset. 

\subsection{Dimension Reduction}

\begin{figure*}[bt]
\vskip 0.2in
\begin{center}
\centerline{\includegraphics[scale=0.9]{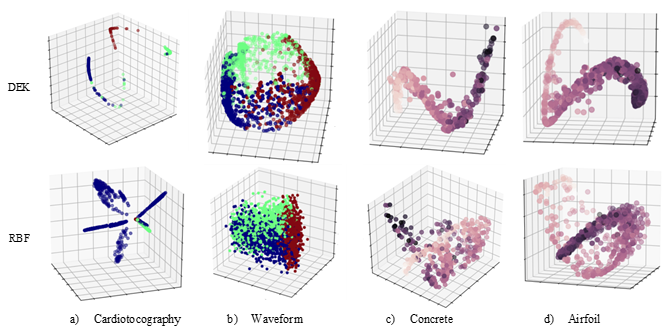}}
\caption{Visualization of Four Datasets in 3D Space}
\label{visualization}
\end{center}
\vskip -0.2in
\end{figure*}

As being a kernel function, a trained DEK can be used with kernel Principal Component Analysis (kPCA) to perform dimension reduction on labeled data. We compare the performance of dimension reduction by kPCA with DEK and kPCA with RBF kernel (hyper-parameter optimized by grid-search with SVM) on two classification datasets (Cardiotocography and Waveform) and two regression datasets (Concrete and Airfoil). Figure \ref{visualization} illustrates the four (testing) sets projected into 3D space by kPCA with a trained DEK and RBF kernel. 

We can observe that, for the two classification datasets, DEK maps the data to a space where classes (represented by nodes’ colors) better match the geographical clusters. For the two regression datasets, the function patterns are clearer in the space mapped by DEK than RBF (the target values are represented by the shades of the nodes -- darker nodes indicate higher target values).

\section{Conclusion}

In this paper, we propose a new learnable kernel that is called DEK to automatically learn an optimized feature space from training data. DEK is represented by a deep neural network that consists of two components: a deep embedding network and a deep kernel network. The integration of these two components in a unified framework is to  maximize the learning power of the deep architecture. The deep embedding network is designed to learn high-level representations; while the deep kernel network is designed to further learn non-linear similarities. Besides the learning capabilities presented by the embedding network and the kernel network, DEK can also integrate deep architectures for embedding learning on unstructured data. DEK can act as a general-purpose kernel function applicable in most supervised learning tasks including identity detection, classification, regression, and dimension reduction. DEK also achieves superior performance on transfer learning for facial recognition compared with frequently used fully connected layers built on pretrained deep networks. We plan to contribute DEK as an open source package on GitHub to promote its usages on different application domains. 

\bibliographystyle{icml2018}
\bibliography{DEK}

\begin{thebibliography}{32}
\providecommand{\natexlab}[1]{#1}
\providecommand{\url}[1]{\texttt{#1}}
\expandafter\ifx\csname urlstyle\endcsname\relax
  \providecommand{\doi}[1]{doi: #1}\else
  \providecommand{\doi}{doi: \begingroup \urlstyle{rm}\Url}\fi

\bibitem[Ayres-de Campos et~al.(2000)Ayres-de Campos, Bernardes, Garrido,
  Marques-de Sa, and Pereira-Leite]{ayres2000sisporto}
Ayres-de Campos, Diogo, Bernardes, Joao, Garrido, Antonio, Marques-de Sa,
  Joaquim, and Pereira-Leite, Luis.
\newblock Sisporto 2.0: a program for automated analysis of cardiotocograms.
\newblock \emph{Journal of Maternal-Fetal Medicine}, 9\penalty0 (5):\penalty0
  311--318, 2000.

\bibitem[Bengio(2012)]{bengio2012deep}
Bengio, Yoshua.
\newblock Deep learning of representations for unsupervised and transfer
  learning.
\newblock In \emph{Proceedings of ICML Workshop on Unsupervised and Transfer
  Learning}, pp.\  17--36, 2012.

\bibitem[Bergstra et~al.(2010)Bergstra, Breuleux, Bastien, Lamblin, Pascanu,
  Desjardins, Turian, Warde-Farley, and Bengio]{bergstra2010theano}
Bergstra, James, Breuleux, Olivier, Bastien, Fr{\'e}d{\'e}ric, Lamblin, Pascal,
  Pascanu, Razvan, Desjardins, Guillaume, Turian, Joseph, Warde-Farley, David,
  and Bengio, Yoshua.
\newblock Theano: A cpu and gpu math compiler in python.
\newblock In \emph{Proc. 9th Python in Science Conf}, pp.\  1--7, 2010.

\bibitem[Bredin(2017)]{bredin2017tristounet}
Bredin, Herv{\'e}.
\newblock Tristounet: triplet loss for speaker turn embedding.
\newblock In \emph{Acoustics, Speech and Signal Processing (ICASSP), 2017 IEEE
  International Conference on}, pp.\  5430--5434. IEEE, 2017.

\bibitem[Breiman et~al.(1984)Breiman, Friedman, Stone, and
  Olshen]{breiman1984classification}
Breiman, Leo, Friedman, Jerome, Stone, Charles~J, and Olshen, Richard~A.
\newblock \emph{Classification and regression trees}.
\newblock CRC press, 1984.

\bibitem[Brooks et~al.(1989)Brooks, Pope, and Marcolini]{brooks1989airfoil}
Brooks, Thomas~F, Pope, D~Stuart, and Marcolini, Michael~A.
\newblock Airfoil self-noise and prediction.
\newblock 1989.

\bibitem[Cummins et~al.(2006)Cummins, Grimaldi, Leonard, and
  Simko]{cummins2006chains}
Cummins, Fred, Grimaldi, Marco, Leonard, Thomas, and Simko, Juraj.
\newblock The chains corpus: Characterizing individual speakers.
\newblock In \emph{Proc of SPECOM}, volume~6, pp.\  431--435. Citeseer, 2006.

\bibitem[Decenci{\`e}re et~al.(2014)Decenci{\`e}re, Zhang, Cazuguel, Lay,
  Cochener, Trone, Gain, Ordonez, Massin, Erginay,
  et~al.]{decenciere2014feedback}
Decenci{\`e}re, Etienne, Zhang, Xiwei, Cazuguel, Guy, Lay, Bruno, Cochener,
  B{\'e}atrice, Trone, Caroline, Gain, Philippe, Ordonez, Richard, Massin,
  Pascale, Erginay, Ali, et~al.
\newblock Feedback on a publicly distributed image database: the messidor
  database.
\newblock \emph{Image Analysis \& Stereology}, 33\penalty0 (3):\penalty0
  231--234, 2014.

\bibitem[Friedman(2002)]{friedman2002stochastic}
Friedman, Jerome~H.
\newblock Stochastic gradient boosting.
\newblock \emph{Computational Statistics \& Data Analysis}, 38\penalty0
  (4):\penalty0 367--378, 2002.

\bibitem[Hermans et~al.(2017)Hermans, Beyer, and Leibe]{hermans2017defense}
Hermans, Alexander, Beyer, Lucas, and Leibe, Bastian.
\newblock In defense of the triplet loss for person re-identification.
\newblock \emph{arXiv preprint arXiv:1703.07737}, 2017.

\bibitem[Hofmann et~al.(2008)Hofmann, Sch{\"o}lkopf, and
  Smola]{hofmann2008kernel}
Hofmann, Thomas, Sch{\"o}lkopf, Bernhard, and Smola, Alexander~J.
\newblock Kernel methods in machine learning.
\newblock \emph{The annals of statistics}, pp.\  1171--1220, 2008.

\bibitem[Hunter(2007)]{hunter2007matplotlib}
Hunter, John~D.
\newblock Matplotlib: A 2d graphics environment.
\newblock \emph{Computing in science \& engineering}, 9\penalty0 (3):\penalty0
  90--95, 2007.

\bibitem[Jiu \& Sahbi(2017)Jiu and Sahbi]{jiu2017nonlinear}
Jiu, Mingyuan and Sahbi, Hichem.
\newblock Nonlinear deep kernel learning for image annotation.
\newblock \emph{IEEE Transactions on Image Processing}, 26\penalty0
  (4):\penalty0 1820--1832, 2017.

\bibitem[Jose et~al.(2013)Jose, Goyal, Aggrwal, and Varma]{jose2013local}
Jose, Cijo, Goyal, Prasoon, Aggrwal, Parv, and Varma, Manik.
\newblock Local deep kernel learning for efficient non-linear svm prediction.
\newblock In \emph{International Conference on Machine Learning}, pp.\
  486--494, 2013.

\bibitem[Liaw et~al.(2002)Liaw, Wiener, et~al.]{liaw2002classification}
Liaw, Andy, Wiener, Matthew, et~al.
\newblock Classification and regression by randomforest.
\newblock \emph{R news}, 2\penalty0 (3):\penalty0 18--22, 2002.

\bibitem[Pan \& Yang(2010)Pan and Yang]{pan2010survey}
Pan, Sinno~Jialin and Yang, Qiang.
\newblock A survey on transfer learning.
\newblock \emph{IEEE Transactions on knowledge and data engineering},
  22\penalty0 (10):\penalty0 1345--1359, 2010.

\bibitem[Pedregosa et~al.(2011)Pedregosa, Varoquaux, Gramfort, Michel, Thirion,
  Grisel, Blondel, Prettenhofer, Weiss, Dubourg, et~al.]{pedregosa2011scikit}
Pedregosa, Fabian, Varoquaux, Ga{\"e}l, Gramfort, Alexandre, Michel, Vincent,
  Thirion, Bertrand, Grisel, Olivier, Blondel, Mathieu, Prettenhofer, Peter,
  Weiss, Ron, Dubourg, Vincent, et~al.
\newblock Scikit-learn: Machine learning in python.
\newblock \emph{Journal of machine learning research}, 12\penalty0
  (Oct):\penalty0 2825--2830, 2011.

\bibitem[Sahbi(2017)]{sahbi2017coarse}
Sahbi, Hichem.
\newblock Coarse-to-fine deep kernel networks.
\newblock In \emph{Proceedings of the IEEE Conference on Computer Vision and
  Pattern Recognition}, pp.\  1131--1139, 2017.

\bibitem[Schmidhuber(2015)]{schmidhuber2015deep}
Schmidhuber, J{\"u}rgen.
\newblock Deep learning in neural networks: An overview.
\newblock \emph{Neural networks}, 61:\penalty0 85--117, 2015.

\bibitem[Schroff et~al.(2015)Schroff, Kalenichenko, and
  Philbin]{schroff2015facenet}
Schroff, Florian, Kalenichenko, Dmitry, and Philbin, James.
\newblock Facenet: A unified embedding for face recognition and clustering.
\newblock In \emph{Proceedings of the IEEE conference on computer vision and
  pattern recognition}, pp.\  815--823, 2015.

\bibitem[Setty et~al.(2013)Setty, Husain, Beham, Gudavalli, Kandasamy, Vaddi,
  Hemadri, Karure, Raju, Rajan, et~al.]{setty2013indian}
Setty, Shankar, Husain, Moula, Beham, Parisa, Gudavalli, Jyothi, Kandasamy,
  Menaka, Vaddi, Radhesyam, Hemadri, Vidyagouri, Karure, JC, Raju, Raja, Rajan,
  B, et~al.
\newblock Indian movie face database: a benchmark for face recognition under
  wide variations.
\newblock In \emph{Computer Vision, Pattern Recognition, Image Processing and
  Graphics (NCVPRIPG), 2013 Fourth National Conference on}, pp.\  1--5. IEEE,
  2013.

\bibitem[Smith et~al.(1988)Smith, Everhart, Dickson, Knowler, and
  Johannes]{smith1988using}
Smith, Jack~W, Everhart, JE, Dickson, WC, Knowler, WC, and Johannes, RS.
\newblock Using the adap learning algorithm to forecast the onset of diabetes
  mellitus.
\newblock In \emph{Proceedings of the Annual Symposium on Computer Application
  in Medical Care}, pp.\  261. American Medical Informatics Association, 1988.

\bibitem[Strobl \& Visweswaran(2013)Strobl and Visweswaran]{strobl2013deep}
Strobl, Eric~V and Visweswaran, Shyam.
\newblock Deep multiple kernel learning.
\newblock In \emph{Machine Learning and Applications (ICMLA), 2013 12th
  International Conference on}, volume~1, pp.\  414--417. IEEE, 2013.

\bibitem[Tang(2013)]{tang2013deep}
Tang, Yichuan.
\newblock Deep learning using linear support vector machines.
\newblock \emph{arXiv preprint arXiv:1306.0239}, 2013.

\bibitem[Tsanas \& Xifara(2012)Tsanas and Xifara]{tsanas2012accurate}
Tsanas, Athanasios and Xifara, Angeliki.
\newblock Accurate quantitative estimation of energy performance of residential
  buildings using statistical machine learning tools.
\newblock \emph{Energy and Buildings}, 49:\penalty0 560--567, 2012.

\bibitem[Vapnik(1999)]{vapnik1999overview}
Vapnik, Vladimir~Naumovich.
\newblock An overview of statistical learning theory.
\newblock \emph{IEEE transactions on neural networks}, 10\penalty0
  (5):\penalty0 988--999, 1999.

\bibitem[Wiering \& Schomaker(2014)Wiering and Schomaker]{wiering2014multi}
Wiering, Marco~A and Schomaker, Lambert~RB.
\newblock Multi-layer support vector machines.
\newblock \emph{Regularization, optimization, kernels, and support vector
  machines}, pp.\  457--476, 2014.

\bibitem[Yeh(1998)]{yeh1998modeling}
Yeh, I-C.
\newblock Modeling of strength of high-performance concrete using artificial
  neural networks.
\newblock \emph{Cement and Concrete research}, 28\penalty0 (12):\penalty0
  1797--1808, 1998.

\bibitem[Zagoruyko \& Komodakis(2015)Zagoruyko and
  Komodakis]{zagoruyko2015learning}
Zagoruyko, Sergey and Komodakis, Nikos.
\newblock Learning to compare image patches via convolutional neural networks.
\newblock In \emph{Computer Vision and Pattern Recognition (CVPR), 2015 IEEE
  Conference on}, pp.\  4353--4361. IEEE, 2015.

\bibitem[Zbontar \& LeCun(2015)Zbontar and LeCun]{zbontar2015computing}
Zbontar, Jure and LeCun, Yann.
\newblock Computing the stereo matching cost with a convolutional neural
  network.
\newblock In \emph{Proceedings of the IEEE conference on computer vision and
  pattern recognition}, pp.\  1592--1599, 2015.

\bibitem[Zhang(1992)]{zhang1992selecting}
Zhang, Jianping.
\newblock Selecting typical instances in instance-based learning.
\newblock In \emph{Machine Learning Proceedings 1992}, pp.\  470--479.
  Elsevier, 1992.

\bibitem[Zhuang et~al.(2011)Zhuang, Tsang, and Hoi]{zhuang2011two}
Zhuang, Jinfeng, Tsang, Ivor~W, and Hoi, Steven~CH.
\newblock Two-layer multiple kernel learning.
\newblock In \emph{Proceedings of the Fourteenth International Conference on
  Artificial Intelligence and Statistics}, pp.\  909--917, 2011.

\end{thebibliography}

\end{document}